# Machine Learning-Based Automated Thermal Comfort Prediction: Integration of Low-Cost Thermal and Visual Cameras for Higher Accuracy

Roshanak Ashrafi[1], Mona Azarbayjani[1], Hamed Tabkhi[1]

[1]UNC Charlotte, Charlotte, NC

ABSTRACT: Recent research is trying to leverage occupants' demand in the building's control loop to consider individuals' well-being and the buildings' energy savings. To that end, a real-time feedback system is needed to provide data about occupants' comfort conditions that can be used to control the building's heating, cooling, and air conditioning (HVAC) system. The emergence of thermal imaging techniques provides an excellent opportunity for contactless data gathering with no interruption in occupant conditions and activities. There is increasing attention to infrared thermal camera usage in public buildings because of their non-invasive quality in reading the human skin temperature. However, the state-of-the-art methods need additional modifications to become more reliable. To capitalize potentials and address some existing limitations, new solutions are required to bring a more holistic view toward non-intrusive thermal scanning by leveraging the benefit of machine learning and image processing. This research implements an automated approach to collect and register simultaneous thermal and visual images and read the facial temperature in different regions. This paper also presents two additional investigations. First, through utilizing IButton wearable thermal sensors on the forehead area, we investigate the reliability of an in-expensive thermal camera (FLIR Lepton) in reading the skin temperature. Second, by studying the false-color version of thermal images, we look into the possibility of non-radiometric thermal images for predicting personalized thermal comfort. The results shows the strong performance of Random Forest and K-Nearest Neighbor prediction algorithms in predicting personalized thermal comfort. In addition, we have found that non-radiometric images can also indicate thermal comfort when the algorithm is trained with larger amounts of data.

PAPER SESSION TRACK: Public Health and Human-Centered Design

## INTRODUCTION

Most people spend over 90% of their time indoors in modern society, causing indoor environmental qualities to significantly influence our health conditions. In this regard, The World Health Organization (WHO) had emphasized the importance of our living place as one of the Social determinants of Health (SDOH), which is defined as "The conditions in which people are born, grow, live, work and age" ("WHO | Social Determinants of Health" 2015). The building sector alone consumes about 40% of global produced energy. Of this amount, the majority is used to provide comfortable interior conditions for building occupants. However, people are still largely dissatisfied with their environmental comfort (Administration and Analysis 2014). Despite several energy-based and hi-tech building systems that are used in recent buildings specially shared open spaces, occupants' comfort level has not been satisfactory (Ashrafi et al. 2019; Mostafavi et al. 2018). This becomes of more importance when we are considering long-term occupancy levels such as office environments (Zarrabi et al. 2018; Armin Amirazar et al. 2018). The recent research regarding the Sick Building Syndrome (SBS) has increased attention to the effect of the built environment on the occupant's health condition, especially at long-term occupancy. It has been proven that temperature and humidity conditions are great contributors to SBS, including fatigue, headache, and susceptibility to cold and flu (Ghaffarianhoseini et al. 2018). Concerning office buildings, lost productivity, decreased performance, and sick absences are causing the most significant losses among environment-related symptoms, which cost businesses around $20 to $70 billion annually. The financial reimbursements of workplace environment improvement were estimated at around $5 to $75 billion annually, resulting in health benefits for more than 15 million workers(Mendell et al. 2002). The Indoor Environmental Quality (IEQ) factors are significant contributors to an office building's comfort and productivity, a combination of thermal, visual (A. Amirazar et al. 2018), acoustics, space layout, and air quality. By understanding the contributing IEQ factors on occupants' comfort, we will improve human well-being and productivity. In this regard, the thermal condition is one of the main contributing factors to SBS, which needs to be studied and improved.
One of the main problems of the buildings' controlling systems is generalized thermal comfort models that have caused discomfort, dissatisfaction, and health-related issues for many building occupants in the indoor space (Shahzad et al. 2017). Most of the current building controlling systems that rely on these explicit pre-defined models of occupant behavior do not correspond to different occupants' actual comfort in the environment. Predicted Mean Vote (PMV) and Adaptive Comfort are the two most common models for controlling indoor thermal conditions. These two models are defined to predict the average thermal comfort of a large population, which has resulted in the uncertainty of these models. As building control systems are currently using these general standards to predict the average thermal comfort of a large population, they are expected to provide comfort for approximately 80% of the building occupants. However, a 10-year- long study of 52,980 occupants in 351 predominantly North American office buildings has shown that only 2% of these buildings provide thermal comfort for 80% of their occupants (Karmann et al., 2018). The large numbers of unhealthy buildings and unsatisfied occupants have made researchers study the validity of the currently used general standards such as PMV. Human attributions such as age, gender, and metabolic rate may affect these preferences. In addition, because the room may be used for different purposes or tasks, occupants and their choices regarding thermal comfort may alter. Along with the physiological characteristics, psychological aspects play a crucial role in a human's mood. Tiredness or emotional status (being happy or angry) and stress level can also influence people's subjective thermal sensation (Hong et al., 2017). This makes it impossible to consider people's differences in thermal preference and the importance of each contributing factor for the individuals by simply relying on general standards(Ruoxi Jia et al. 2018).

Recent research is developing the notion of personalized comfort by attempting to leverage occupants' demand in the control loop of buildings to consider the well-being of each individual based on their personal physiological properties. Therefore, a real-time feedback system is needed to provide data about occupants' physiological conditions that can control the building's heating, cooling, and air conditioning (HVAC) system. Personalized comfort is a recent concept in the building design area that provides comfortable conditions for each occupant based on their preferences. The innovations in environmental data gathering have provided an excellent opportunity to collect large amounts of information from the buildings' occupants, which can be studied to improve a building's control conditions. In this regard, the emergence of thermal imaging techniques makes contactless data gathering possible without any interruption in occupant conditions and activities.

In this research, we are looking into the possibility of using low-cost thermal cameras as a cost-effective vision-based method for automated data gathering of occupants' thermal conditions. In addition, this research creates a fully automated platform for a more precise reading from larger distances, which makes it an excellent fit for real-time applications in the actual world. This is completed by leveraging simple visual(RGB) and thermal cameras to create a multimodal sensing platform. Through this integrated system, we will use visual cameras to localize facial areas (e.g., forehead, cheeks, nose) while using thermal cameras to measure the thermal values of those areas, thus enhancing the accuracy and robustness of sensing and measurement. These features would make this approach optimal to be used in multi-occupancy spaces such as office environments. The study has two main contributions:

First, an automated personalized thermal comfort prediction model is developed by integrating low-cost thermal and visual cameras.

Second, the prediction accuracy of different variables, including thermal infrared skin temperature, thermal image pixel intensity, and wearable sensors, are compared with each other.

The rest of this paper has three main sections in section 1. Literature Review looks into the conventional comfort models and the current state-of-the-art research on personalized thermal comfort. Section 2 explains our data collection and analysis methodology, and the results are presented and analyzed in Section3.

## 1.0 Literature Review

This section will summarize current thermal comfort models and alternative approaches to taking into account the preferences of building occupants. Finally, we will discuss infrared imaging as a non-contact method for collecting human-centric data and successful research.

The Predicted Mean Vote (PMV) is a widely used model for assessing thermal comfort that Fanger developed in 1960 to represent the average thermal sensation vote of a large group of people(Cheng, Niu, and Gao 2012). This model was created based on the difference between generated heat and released heat from the human body and its correlation with the subjective perception of comfort. Since the PMV model was developed in a chamber setting within an air-conditioned space, the results are expected to differ in natural settings and naturally ventilated buildings. Previous research backs up this model for higher quality performance results in natural ventilation buildings(Rupp, Vásquez, and Lamberts 2015)(Humphreys and Fergus Nicol 2002). A PMV model was initially developed to predict the thermal sensation of groups of people. However, the prediction accuracy for groups of people was not acceptable in several studies(Cheung et al. 2019). The PMV model's only acceptable prediction was in neutral conditions within the range of 0.25, as bias was shown in both sides of the cool and hot sensations, with poorer performance on the cool side (Humphreys and Fergus Nicol 2002; Cheung et al. 2019). In several studies, the PMV factor performed poorly in both individual and group level predictions of thermal sensation compared to observed thermal sensation. The leading cause of this inaccuracy is a variety of individual differences resulting in different thermal preferences that were not considered in the PMV calculation. The PMV model's low accuracy raises concerns about using it to control our buildings.

The Human-in-the-loop (HITL) concept has redefined the relationship between humans and their surrounding environments controlling systems. To achieve a high-performance building throughout the operation phase, embracing subjective human aspects in the control loop is necessary. Providing the desired temperature set point to minimize discomfort among all occupants is an important yet challenging problem. HITL methods enhance building management performance to take advantage of users' feedback and receive an adaptive model at each iteration. The Internet of Things (IoT) is a recent technology that facilitates communication between gadgets and building inhabitants. Buildings with IoT technology employ a real-time monitoring system to make this task viable. IoT-based systems and HITL techniques enable device-to-device connectivity and data exchange for sensing, actuation, and control (Ray 2018) An efficient IoT system requires data-gathering devices that offer real-time feedback to the controlling loop. Sensors can collect data on both the environment and the occupants' physiological state. These data collection devices can either be intrusive or non-intrusive to building inhabitants. The rapid advancements in environmental data collection have created an invaluable opportunity for amassing large amounts of data that may be evaluated to improve the quality of our interior environment. The majority of current HITL research uses occupancy-based models that rely entirely on occupancy detection techniques such as motion, visual representation, location, and the usage of devices to create schedules and regulate thermal conditions. Other research uses direct feedback from occupants to control conditions using the voting and physiological sensing systems(Jung and Jazizadeh 2018). In participatory sensing systems, thermal scale preferences quantify comfort(Jazizadeh, Marin, and Becerik-Gerber 2013; Erickson and Cerpa n.d.). These intrusive voting systems require constant feedback from the occupants. Use wristbands to collect physiological and environmental data to predict each occupant's comfort(Dai et al. 2017; Choi and Yeom 2017), including skin temperature (Dai et al. 2017), heart rate (Choi and Yeom 2017), or both (Liu et al. 2019). These data collection methods are also considered intrusive because the sensor devices must contact the human skin all day. This data-gathering obstacle shows the extent to which considering contactless, non-intrusive approaches can be beneficial for obtaining personal physiological data for each occupant.

The emergence of thermal imaging techniques provides an excellent opportunity for contactless data gathering with no interruption in occupant conditions and activities. In this research, we are looking into the possibility of a non-contact vision-based method from a distance for gathering data on occupants' thermal conditions. In an attempt to create a non-invasive data gathering approach, infrared sensors were installed on eyeglasses to gather the temperature of the front face, cheeks, nose, and ears, which increased accuracy to % 82.8 for the prediction of uncomfortable conditions(Ghahramani et al. 2018). Although this device is not in direct contact with the skin, it cannot be considered a non-intrusive approach as this is still a wearable device. Infrared thermal cameras can replace these infrared sensors because they provide more non-intrusiveness through thermal imaging techniques. The infrared cameras can be installed far from the occupant and capture the skin temperature by reading the pixel values of the desired regions. By proving the feasibility of this technique with the accuracy of 94% -95 % when using FLIR A655sc (Ranjan and Scott 2016), infrared thermography has been verified as an accurate non-intrusive approach. A real-time feedback system using FlirA35 thermal camera was developed in 2018 and analyzed face temperature and occupants' position(Metzmacher et al. 2018). Researchers could replace the previously mentioned cameras with a lower-cost and smaller infrared camera with an acceptable accuracy of 85% for predicting the skin temperature compared to the expensive high-resolution cameras(Li, Menassa, and Kamat 2018). Researchers have also compared different facial feature detection algorithms to check the accuracy of each approach in detecting the regions of interest (ROIs) (Aryal and Becerik-Gerber 2019). One of the recent studies in this area has compared the accuracy of using three different sensor types, including air temperature sensors, skin temperature with a wristband, and face temperature through thermal imaging. This study highlights the slight improvement in accuracy by adding physiological sensors to the environmental sensors while questioning the efficiency of using physiological sensors due to this small accuracy increase (%3-%4)(Aryal and Becerik-Gerber 2019) In another recent study in this area, Li et al. successfully monitored and recorded two occupants' skin temperature simultaneously with two thermal camera nodes. In contrast, each camera captured some parts of the faces (Li, Menassa, and Kamat 2018)

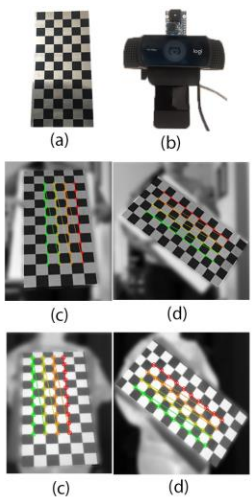

**Figure 1.** Camera Calibration (a) constructed checkerboard, (b) Dual Camera system, (c,d)sample detected checkerboard corners by both cameras

## 2.0 Methodology

This section explains the primary data analysis platform setup through the dual-camera system followed by the data collection method explanation.

Infrared skin temperature is measured through our combined dual-camera system, including a FLIR Lepton 3.5 thermal infrared camera and a Logitech C922 RGB camera (Figure1). The resolution of the thermal camera is 160*120 pixels with the radiometric accuracy of ±5°C and a measurement resolution of 0.1 °C, and the resolution of the Logitech camera is 1,280*960 pixels. Initially, a checkerboard registration approach was conducted to calibrate the cameras' intrinsic factors together, as explained by Li 2019 (Li, Menassa, and Kamat 2019a). As shown in Figure 1, a checkerboard was constructed using an aluminum sheet and vinyl polymer material with different heating values, which were heated to be detectable by the thermal camera with the checker pattern. To generate an intrinsic registration matrix between the two cameras, the checkerboard pattern must simultaneously be caught by both cameras in various angles and orientations. After calibrating the two cameras together, thermal and RGB images captured simultaneously can read the desired facial Regions of Interest (ROI). This step is a very important and sensitive task, since wrong calibrated images will result in assigning the wrong temperature to the ROI, especially in the facial area that include adjacent areas with high temperature difference. The thermal reading method is shown in Figure 2. And explained in detail as follows:

(a) The facial area is detected by the RGB camera and cropped in both images. After working with different face detection algorithms, we used a Dlib-based face recognition model to crop the facial area. This model is based on a 29 convolutional layer in Residual Networks ResNet (He et al. 2016).
(b) Two masked images are created from thermal and RGB frames to precisely calibrate the two images together.
(c) The Homography matrix is defined by using the oriented fast and rotated BRIEF (ORB) characteristics between the two masked images as previously performed by Negishi et al. (Negishi et al. 2020) for detecting the respiratory rate in medical applications.

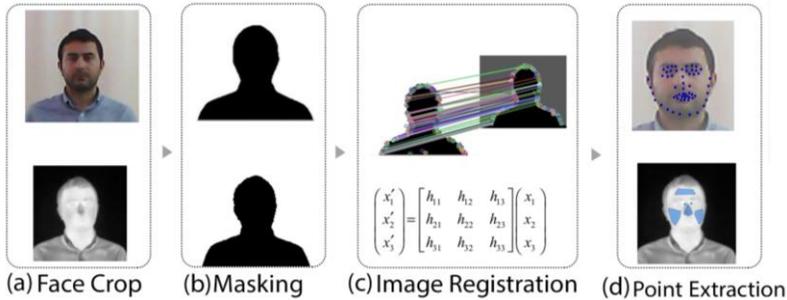

**Figure 2.** Image Registration and Thermal Reading

(d) Facial landmarks are defined from the RGB images, and desired ROIs are calculated based on them. Sixty-eight facial landmark coordinates are detected in each RGB image and are transferred to the thermal image by utilizing the developed homography matrix in the previous step.
(e) Based on the defined ROIs, we can read either the skin temperature or the pixel intensity value based on thermal image type, 16bit radiometric images, or 8-bit non-radiometric and false-colored images accordingly.

## 2.1. Data Collection

The experiment is designed in two separate settings to ensure a comprehensive understanding of different aspects and contributing variables in prediction accuracy. Under IRB183845, the Office of Research Protections and Integrity has provided its clearance to this study. The data collection process began in January of 2021 and continued until April of 2021. The subjects are five healthy individuals aged 33-43 years old and all students. Before the testing, we ensured that the subjects were not suffering from any thermoregulatory disorders such as heat intolerance, colds, flu, or infections. In addition, the participants were instructed not to wear any makeup or facial moisturizer and to remove their glasses for the recording sessions. All the participants were dressed in a dark-colored long-sleeved shirt and pants.

### 2.1.1 Experiment 1

The preliminary experiment is designed to investigate the thermal preference prediction accuracy when utilizing two variables: 1) facial skin temperature readings by a radiometric enabled thermal camera and 2)pixel intensity data of converted non-radiometric thermal images. Personal comfort models are developed based on each individual's physiological or behavioral data through time in diverse thermal conditions for each occupant. While we do not need many subjects, each subject must be studied under several thermal conditions to provide as much data as possible for training the algorithms.

The experiment chamber is a temperature-controlled room, as is shown in Figure 3. After taking the informed consent, the participant's age, gender, height, and weight were recorded. The participants had entered the test room and stayed in a seated position for 30 minutes before the test, so their metabolic rate reached a stable state, and any influence of the prior outdoor temperature was eliminated. The test sessions began at 22°C (71.6°F) and lasted 60 minutes, consisting of six sessions of 10 minutes static and transient conditions. The temperature was kept constant for 10 minutes and then was raised to 2°C (~5°F) higher for the next 10 minutes. This pattern was repeated three times to reach the final temperature of 28°C(~82°F). This technique generated both transient and static thermal conditions. The environmental sensors are HOBOProv2 temperature/relative humidity data logger sensors, mounted on a pole beside the subject's station at a 0.5-meter distance and at three different heights (0.1, 1.1, and 1.7 meters) to record the temperature and relative humidity.

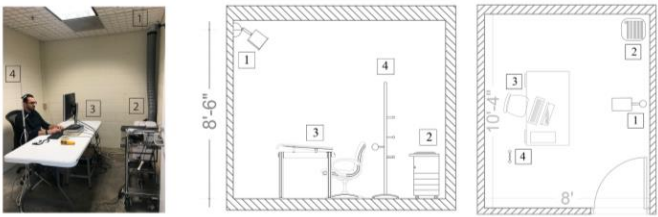

**Figure 3**. Study Chamber (1)Dual Camera (2)Air Conditioning (4) Data Logger (4)Environmental Sensors

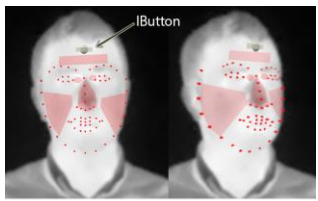

**Figure 4** . ROI Extraction from landmars and IButton placement

The dual-camera system is installed on the monitor in front of the user with a 1-meter (3.2ft) distance from the subject. Every five seconds, synchronized images were captured by both thermal and RGB cameras, and the skin temperature data was recorded using a thermal camera and an IButton DS1923 sensor attached to the highest point of the forehead (Figure 4). The users' subjective thermal sensation and preference were also recorded every two minutes based on their answer to the questions of "What is your current thermal sensation? (Cold, cool, slightly cool, neutral, slightly warm, warm, and hot) and "How do you prefer your thermal environment to change?" (Warmer, slightly warmer, no change, slightly cooler, colder)

### 2.1.2 Experiment 2

The second set of experiments are designed to investigate the prediction accuracy of non-radiometric false-color thermal images. The experiments were conducted in the same room setting, with three subjects participating individually. The main objective of this phase was to look into the correlation of different facial areas' thermal intensity with the air temperature. The thermostat temperature was increased from 21°C (~70°F) to 28°C(~83°F) in 90 minutes with a fixed rate. The thermostat setpoint was increased by 1°C every six minutes; however, the air temperature sensors show variation, some at an increasing rate. The thermal sensors are placed as the previous session to record the temperature and relative humidity. In this session, the camera system was mounted on the wall in front of the subject with a 3 Meter (9.8ft) distance from the camera, as is shown in Figure 4. To increase the number of data points, the capturing intervals were decreased to one frame per second. The subjective thermal sensation and preference of the users are indicated as in the previous experiment, every three minutes.

## 3. Results and Discussion

### 3.1. Experiment 1

The main objective of this experiment is to compare the prediction accuracy of skin temperature from higher accuracy radiometric images with the thermal intensity from 8bit false-color thermal images. As mentioned in the literature review section, recent previous research has shown that facial skin temperature extracted from radiometric images can predict the subjective thermal preference of the user through the creation of personal thermal comfort models (Li, Menassa, and Kamat 2018; 2019b; Aryal and Becerik-Gerber 2019; Cosma and Simha 2019). However, many of the currently available thermal cameras do not include the radiometric option, and therefore, the skin temperature data for each pixel would not be available. A total of 720 data frames were captured for each subject, with intervals of 5 seconds in a 60-minute experiment.

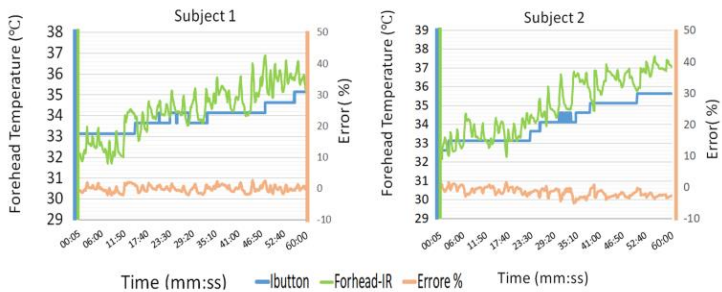

**Figure 5.** Forehead temperature difference between IButton &Thermal Camera

Initially we looked into the accuracy of the thermal camera readings by comparing it to a more accurate wearable sensor. We have compared the IButton DS1923 sensor's temperature readings and extracted skin thermal camera in the forehead area to validate our thermal camera setting and accuracy. IButton sensors have an accuracy range of ±0.5°Ca and were previously shown to have acceptable performance and accuracy for reading the skin temperature (Liu et al. 2019). The results are presented in Figure 5, which shows both temperature readings and the difference between these two readings. As the error percentage is less than 2% for both subjects, we can conclude the reliability of our skin temperature extraction method and the thermal camera's accuracy.

Furthermore, we look into the changes in room temperature, skin temperature, and lower quality images' thermal intensity in one graph for both subjects. In this regard, the initially captured 16bit and radiometric frames were converted to 8bit grayscale RGB888 images to study both sets of data frames. The extracted data first needed to be cleaned and filtered. As shown in Figure 6. although both skin temperature and pixel intensity increase as we increase the room temperature with a reliable pattern, there are some inconsistencies in the thermal camera readings. This change in thermal measurements is due to the camera's Fast Field Correction (FFC) action, which is executed every 3 minutes to recalibrate the camera. We have removed the outliers and eliminated this effect by removing the outliers and adding a moving average filter with a 5 data points period. The results of filtered data are also presented in Figure 5.

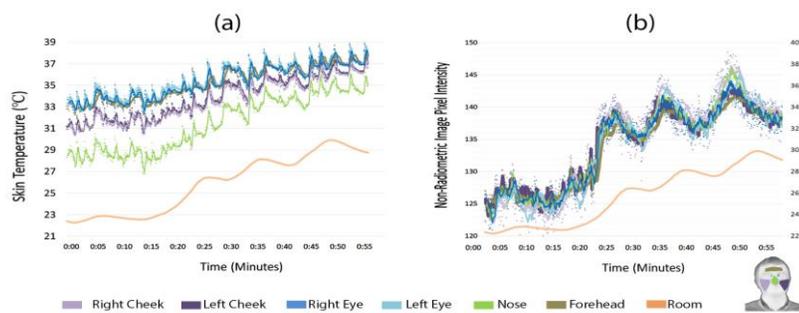

**Figure 6.** Raw and processed (a)skin temperature and (b)pixel intensity

The correlation between room temperature and skin temperature values and thermal intensity is calculated based on Pearson Correlation Coefficient to better understand the changing pattern in the heating transient condition. The Pearson correlation coefficient indicates the strength and direction of a linear relationship between variables and is relevant for our calculations. As presented in Table 1., the nose area has the highest correlation with the room temperature for both subjects, which is 0.97 for subject1 and 0.96 for subject 2. The next highest correlation coefficient for both subjects are the cheeks areas, which are interestingly the same number for both sides. We need to mention that the cheeks do not show the same temperature patterns under all circumstances. This homogeneous temperature may result from running the tests in a fully controlled test chamber. The forehead and eye area have the lowest correlation with the environmental temperature, which proves the applicability of this facial area as an indicator of core body temperature, as the environmental properties have less influence on this area. This result is in line with the previous findings from (Silawan et al. 2018) on the correlation of different facial regions with the environmental temperature. The same table also presents the calculated correlation from the second set of image frames with false color and lower quality 8bit format. The correlation numbers with the room temperature in the lower quality images' pixel intensity are lower. They do not follow the same pattern of skin temperature, which might be expected. However, the nose area still has the highest correlation with the room temperature among other facial regions.

| | | Nose | Forehead | Right Cheek | Left Cheek | Right Eye | Left Eye |
|---|---|---|---|---|---|---|---|
| S1 | Skin Temperature | 0.97 | 0.94 | 0.95 | 0.95 | 0.92 | 0.91 |
| | Pixel Intensity | 0.90 | 0.91 | 0.83 | 0.85 | 0.89 | 0.82 |
| S2 | Skin Temperature | 0.96 | 0.92 | 0.94 | 0.94 | 0.89 | 0.91 |
| | Pixel Intensity | 0.89 | 0.88 | 0.83 | 0.89 | 0.87 | 0.76 |

**Table 1.** Correlation of different facial regions with the room temperature in both image types

Furthermore, the thermal preference prediction accuracy of three machine learning algorithms is calculated and compared. For each participant, 30 subjective thermal sensation and thermal preference data were recorded. As previously studied by researchers, thermal preference is a better indicator of thermal comfort, so in this paper, we are not working with subjective thermal preference data. The responses were divided into three categories, with "Slightly Warmer" and "Warmer" assigned to a "Warmer" category and "Slightly Cooler" and "Colder" assigned to a "Cooler" category. Since 720 thermal frames and 30 subjective votes were recorded, each subjective vote was assigned to the thermal frames between two voting sessions (Every 2 Minutes). To assess the efficacy of machine learning algorithms for predicting thermal comfort, we trained three previously shown successful algorithms in predicting personalized thermal comfort from the literature. Random Forest, Support Vector Machine (SVM), and K-Nearest Neighbor (KNN) were trained and tested with each subject's personalized physiological and subjective data. Table2 shows the average accuracy of these three algorithms for all three subjects. In addition, the precision for each class category is presented. The Random Forest produces the highest prediction results for both subjects in both radiometric and non-radiometric images, which are (0.93, 0.79) for subject1 and (0.90, 0.81) for subject 2. The SVM algorithm had predicted better than KNN (n=6) for the first subject but worse for the second one. Therefore we cannot conclude which algorithm would be a better choice from these two. We can also see that all algorithms have better performance for both subjects when working with radiometric images and skin temperature. However, the performance of pixel intensity is still acceptable, primarily when it was used to train the Random Forest algorithm (0.79 and 0,81). For the next experiment, we will study the performance of these three algorithms with lower quality non-radiometric images from a larger distance.

| ID | | Random Forest | | | | KNN | | | | SVM | | | |
|---|---|---|---|---|---|---|---|---|---|---|---|---|---|
| | | Accuracy | Precision | | | Accuracy | Precision | | | Accuracy | Precision | | |
| | | | Cooler | No Change | Warmer | | Cooler | No Change | Warmer | | Cooler | No Change | Warmer |
| S1 | Skin Temp. | 0.93 | 0.89 | 0.89 | 1.00 | 0.87 | 0.81 | 0.84 | 1.0 | 0.82 | 0.86 | 0.70 | 0.90 |
| | Intensity | 0.79 | 0.97 | 0.62 | 0.71 | 0.73 | 0.93 | 0.57 | 0.70 | 0.60 | 0.96 | 0.56 | 0.00 |
| S2 | Skin Temp. | 0.90 | 0.87 | 0.85 | 1.00 | 0.78 | 0.77 4 | 0.68 | 0.89 | 0.86 | 0.88 | 0.78 | 0.95 |
| | Intensity | 0.81 | 0.85 | 0.68 | 0.89 | 0.78 | 0.80 | 0.64 | 0.89 | 0.78 | 0.68 | 1.0 | 0.67 |

**Table 2.** Prediction accuracy of three selected prediction algorithms in both image types

### 3.1. Experiment 2

In the second set of experiments, we have recorded the frames without the radiometric option to have a detailed look at the prediction accuracy of lower quality and false-colored 8bit thermal images from a 3-meter distance. We increased the number of data frames to compensate for the accuracy decrease due to this conversion and increased camera distance. The data reading interval was decreased to 1 second, which has provided us with approximately 5000 data points for each subject. The intervals of recording the thermal preference were increased to three minutes due to the previous subjects' feedback. For each participant, 30 subjective thermal votes were recorded. In addition, we have changed the thermal preference categorizing patterns to 4 levels and divided the "cooler" preference into two categories of "slightly cooler" and "colder." The lowest starting temperature was around 21°C (70°F) to ensure the subjects were not exposed to extreme cold conditions for health considerations; the cold temperature duration is less than the hot segment and does not need to be divided into categories. Since the distance to the camera has increased and we are using lower quality images, we have excluded the eye are from the calculations to avoid accuracy due to the low number of pixels in those areas.

Figure.7 presents the change in the room temperature and pixel intensity of the selected areas with a polynomial regression of 6th degrees applied to them, along with the subjective thermal vote of the participants. As shown in Figure.7, the subjective thermal preference of the participants starts with "slightly warmer" as they were subjected to an approximate temperature of 21°C (70°F) for 20 minutes before the start of data recording. The figure shows that all three subjects have experienced the four thermal preferences, while "No Change" is the majority of votes for all. It is also demonstrated that the changing pattern in the skin temperature is different for each subject. The facial areas in subject 3 have a closer temperature together.

On the other hand, subject 5 has a similar temperature in the cheeks area, while the forehead and nose area's temperature is more similar and different from the cheeks. Another interesting point about cheeks' temperature in subject5 is their considerable correlation with the subject's thermal preference. The subject's thermal preference was "No change" at the beginning of the experiment, but after 10 minutes, as the cheeks temperature decreases due to the cold weather, the subject would prefer a slightly warmer environment.

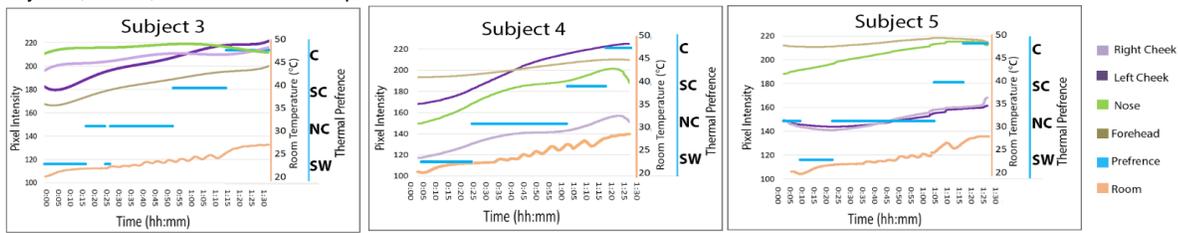

**Figure 7.** Changes in room temperature, pixel intensity, and thermal preference (C: colder, SC: slightly colder, NC: no change, SW: slightly warmer)

Table 2. shows the Average Accuracy for all three subjects by the selected algorithms. In addition, the precision for each class category is presented. The SVM algorithm shows the lowest prediction results for all three subjects, similar to the previous results (0.82, 0.85, 0.86). On the other hand, the Random forest and KNN algorithms show approximately close results in both the Average Accuracy and Precision in the prediction of each class. The precision results for both of these algorithm are better in predicting cold preference than warmer preference. However, the SVM algorithm had performed better in the warm preference class.

| ID | Random Forest | | | | | KNN | | | | | SVM | | | | |
|---|---|---|---|---|---|---|---|---|---|---|---|---|---|---|---|
| | Accuracy | Precision | | | | Accuracy | Precision | | | | Accuracy | Precision | | | |
| | | Colder | Slightly Cooler | No Change | Warmer | | Colder | Slightly Cooler | No Change | Warmer | | Colder | Slightly Cooler | No Change | Warmer |
| S3 | 0.95 | 0.98 | 0.92 | 0.95 | 0.96 | 0.95 | 0.98 | 0.92 | 0.95 | 0.94 | 0.85 | 0.94 | 0.73 | 0.81 | 0.92 |
| S4 | 0.95 | 0.97 | 0.89 | 0.98 | 0.97 | 0.96 | 0.96 | 0.93 | 0.98 | 0.97 | 0.82 | 0.83 | 0.61 | 0.90 | 0.92 |
| S5 | 0.9 | 0.99 | 1.00 | 0.98 | 0.94 | 0.95 | 0.99 | 0.99 | 0.98 | 0.86 | 0.86 | 0.92 | 0.61 | 0.67 | 1.00 |

**Table 3.** Prediction accuracy of three selected prediction algorithms for all three subjects

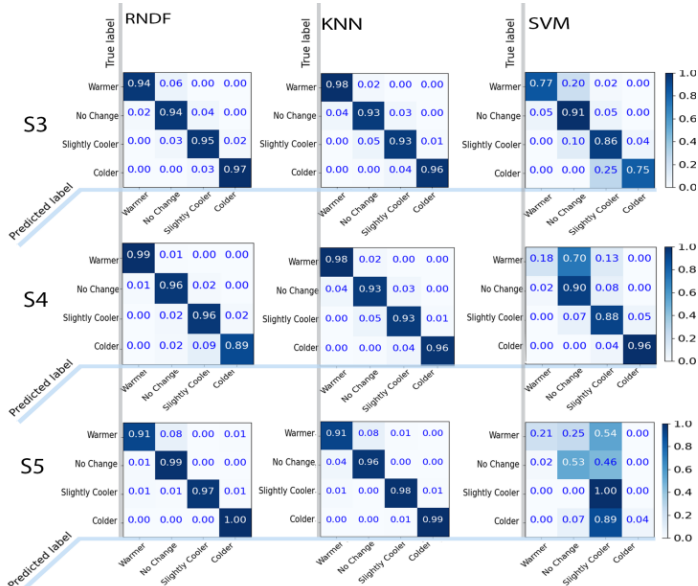

**Figure 8.** Confusion matrices for three prediction algorithms

The confusion matrices for all three subjects are presented in Figure.8 to comprehensively analyze the three algorithms' performance in predicting different thermal preference classes. This figure shows that the SVM classifier has the lowest number of True predictions and the highest amounts of False ones in all class categories. The performance of this algorithm is deficient, especially in predicting warmer preferences. In addition, the number of false predictions is much higher than the other two algorithms in several classes.

The performance of Random Forest and KNN algorithms are primarily close together, with some slight differences. The only notable difference is predicting colder preference in Subject 4, 0.89% for Random Forest and 96% for the KNN algorithm.

**CONCLUSION**

An automated infrared thermal reading platform was studied to predict the personalized thermal preference in heating transient conditions. The highlight of this approach is the automatic calibration of thermal and RGB images without manual registration or knowing the subjects' distance from the camera. The tests were conducted in two sessions to compare the feasibility of utilizing non-radiometric images instead of higher-quality radiometric images. It was found that although the prediction performance of radiometric enabled images was superior to non-radiometric ones, by increasing the number of data frames, we can obtain very high accuracy predictions from lower quality images. Another finding of this study is the better performance of Random Forest and KNN to SVM prediction algorithm, which is also in line with the prior research in this area. Therefore we do not recommend utilizing SVM algorithms for personalized thermal comfort prediction.

This study also has some limitations that need to be addressed in our current and future tests on personalized thermal comfort prediction. Although the number of subjects does not need to be many in a personalized prediction algorithm, we still need to perform this research with more subjects to study the prediction performance in more diverse physiological properties. In this set of experiments, we had not recorded the air velocity. We will also add air velocity to the calculations and study its influence on thermal sensation and preference for the next experiments. Furthermore, the subject's distance to the camera was not changing at the time of the experiment. Also, the subject's head positions were mainly the same during the experiment: full frontal face with minor amounts of yaw or pitching at times. Research has shown that the distance to the camera and angle with the thermal sensor influences the infrared thermal readings. It is important to perform the experiment at several other distances from the camera and at different head positions. We are currently conducting another set of experiments, which will be presented in our future publications. Finally, although research in test chambers and transient conditions is an excellent approach for gathering large amounts of data in less time, it may not be as realistic as the data collected through time in a natural office setting. We are also working on performing these tests in an actual office building set for our future research.

**ACKNOWLEDGEMENTS**

This publication resulted from research supported by The School of Data Science at UNC Charlotte (Data Science Initiative award), and National Science Foundation (Award number NSF2104223).